\documentclass[a4paper, oneside, twocolumn, notitlepage, 10pt]{extarticle_ecoc}
\usepackage{ecoc}

\def\ve#1{{\mathchoice{\mbox{\boldmath$\displaystyle #1$}}%
{\mbox{\boldmath$\textstyle #1$}}%
{\mbox{\boldmath$\scriptstyle #1$}}%
{\mbox{\boldmath$\scriptscriptstyle #1$}}}}
\def\j{\mathrm{j}}

\newcommand{\Dkl}{\mathrm{D_{KL}}}

\newcommand{\argmaxst}[3]{%
  &\arg\max_{#1} #2 \\
  &\text{subject to} \quad #3
}

\begin{document}
\selectlanguage{english}    


\title{
Neural Probabilistic Shaping: Joint Distribution Learning for Optical Fiber Communications
}%


\author{
    Mohammad Taha Askari\textsuperscript{(1)}, Lutz Lampe\textsuperscript{(1)},
    Amirhossein Ghazisaeidi\textsuperscript{(2)}
}

\maketitle                  


\begin{strip}
    \begin{author_descr}

        \textsuperscript{(1)} Department of Electrical and Computer Engineering, University of British Columbia, Vancouver, BC V6T 1Z4, Canada,
        \textcolor{blue}{\uline{mohammadtaha@ece.ubc.ca}}
        
        \textsuperscript{(2)} Nokia Bell Labs, 12 rue Jean Bart, 91300 Massy, France

    \end{author_descr}
\end{strip}

\renewcommand\footnotemark{}
\renewcommand\footnoterule{}


\begin{strip}
    \begin{ecoc_abstract}
        We present an autoregressive end-to-end learning approach for probabilistic shaping on nonlinear fiber channels. Our proposed scheme learns the joint symbol distribution and provides a 0.3-bits/2D achievable information rate gain over an optimized marginal distribution for dual-polarized 64-QAM transmission over a single-span 205~km link. \textcopyright2025 The Author(s)
    \end{ecoc_abstract}
\end{strip}




\section{Introduction}

Probabilistic shaping (PS) has been widely adopted in communication systems to approach channel capacity by optimizing the symbol distribution \cite{bocherer2015bandwidth}. While PS is known to provide linear shaping gains over the additive white Gaussian noise (AWGN) channel, it can also yield additional gains in optical fiber systems when the distribution is carefully tailored to account for channel nonlinearities and memory \cite{askari2024probabilistic}. The work in \cite{dar2014shaping} showed that shaping over symbol sequences under an energy constraint can provide nonlinear shaping gains, with the optimal block length closely tied to channel memory. Furthermore, finite blocklength shaping methods such as enumerative sphere shaping (ESS) \cite{amari2019introducing} have been shown to mitigate nonlinear interference noise (NLIN) \cite{fehenberger2020analysis}. Subsequent studies revealed that 
differences in temporal symbol structures, i.e., joint distributions, can significantly affect NLIN characteristics \cite{askari2024probabilistic, askari2023probabilistic}.

Sequence selection is an indirect method for optimizing joint symbol distributions using rejection sampling \cite{secondini2022new}. This approach generates multiple symbol sequences 
and selects those that minimize NLIN using a nonlinearity-aware metric  \cite{askari2024probabilistic, civelli2024sequence}. Although effective in shaping joint distributions, the method does not ensure optimality   because the quality of candidate sequences is uncontrolled, and the selection metrics may not fully capture channel impairments or digital signal processing (DSP) effects. Consequently, numerous candidate sequences are required to achieve meaningful performance gains,  increasing computational complexity and incurring additional rate loss 
\cite{civelli2024cost}.

Autoencoder (AE) neural networks can directly optimize PS by modeling the transmitter, channel, and receiver as a single differentiable system  for end-to-end (E2E) optimization. In \cite{stark2019joint}, a symbol-wise AE generates logits for a marginal distribution and uses the Gumbel-softmax trick \cite{jang2016categorical} for differentiable sampling 
and the cross-entropy loss to maximize mutual information over the AWGN channel. Later, \cite{ait2020joint} introduced a bit-wise AE that jointly optimizes symbol distribution, bit labeling, and demapping to maximize bit-metric information using log-likelihood ratios (LLRs). E2E learning has also been applied to both coherent and non-coherent optical fiber systems \cite{karanov2018end, rode2023end, neskorniuk2022model}. However, most existing work focuses on the marginal symbol distribution, which is suboptimal in channels with significant nonlinearity and memory. Only a few studies have attempted to incorporate memory effects: one using a learned pre-distortion filter while still relying on marginal-distribution shaping \cite{neskorniuk2022memory}, and another explores joint-distribution learning over two-symbol blocks in a four-dimensional constellation, though without scalability to longer sequences \cite{liu2022probabilistic}.

This paper introduces a new E2E learning framework that directly optimizes the joint distribution of symbol sequences for nonlinear fiber channels. Its core is a recurrent neural network (RNN) to generate conditional symbol distributions using  Gumbel-softmax sampling for  gradient-based training. We refer to this architecture as neural PS (NPS). 
The framework supports any differentiable channel model and DSP module and can be implemented in a distribution matcher without the 
rate loss inherent to sequence selection. 
Numerical results show that joint-distribution learning with NPS improves achievable information rate (AIR) and mitigates NLIN more effectively than marginal-distribution shaping, finite-blocklength shaping, or sequence selection.

\section{Problem Statement}
To enable sequence-level PS, we aim to learn a joint distribution over transmitted symbols that maximizes the AIR under bit-metric decoding (BMD). Let $\ve{x} = (x_1, \dots, x_L)$ and $\ve{y} = (y_1, \dots, y_L)$ denote the transmitted and received sequences of length $L$, respectively. Each transmitted symbol $x_t$ is drawn from an $M$-ary constellation set ${\cal C} = \{c_1, \dots, c_M\}$ and is associated with a binary label  $\ve{b}_t = (b_t^{(1)}, \dots, b_t^{(m)})$, where $m=\log_2(M)$.

The learning objective is to optimize the joint distribution $p(\ve{x})$ to maximize the AIR, which for BMD can be written as \cite{ ait2020joint}
\begin{align}
\label{eq:BMD_rate}
\argmaxst{p(\ve{x})}{\sum_{t=1}^{L} \bigg[\underbrace{H(\ve{b}_t) - \sum_{i=1}^{m} H(b_t^{(i)}|{\ve{y}})}_{R_t} \bigg]}{\sum_{\ve{x}}p(\ve{x})\|\ve{x}\|^2=1 \label{eq:constraint}},
\end{align}
where $H(\cdot)$ denotes entropy, $R_t$ is the AIR contribution from the $t$-th symbol, and the constraint enforces unit average transmit power. 
Since directly optimizing the $M^L$-dimensional joint distribution is intractable, we seek a structured and trainable parameterization. In the next sections, we present the NPS architecture designed to efficiently model this distribution while enabling E2E optimization through gradient-based learning.

\section{Neural Probabilistic Shaping Encoder}
\label{sec:encoder}

\begin{figure}[t!]
    \hspace*{-5mm}\includegraphics[width=1.1\linewidth]{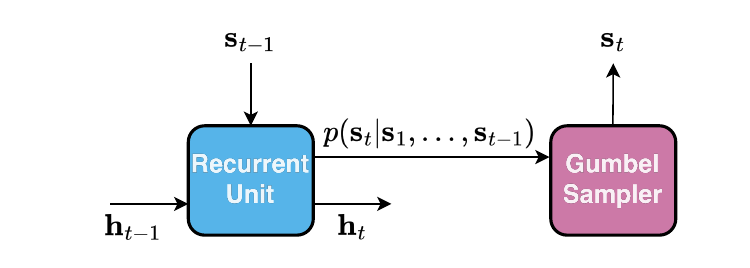}
    \caption{NPS encoder block diagram at time step $t$. $\ve{h}_t$ is the recurrent context vector, and $p(\ve{s}_t \mid \ve{s}_{1},\dots,\ve{s}_{t-1})$ is the predicted conditional distribution. The one-hot vector $\ve{s}_t$, representing the transmitted symbol $x_t$, is sampled from this distribution using the Gumbel-softmax trick.}
    \label{fig:encoder_blockdiagram}
\end{figure}


The core idea is to use an encoder that parameterizes the factorized joint distribution $p(\ve{x})$ using an autoregressive model. At each time step $t$, the encoder generates the logits for the conditional distribution $p(x_t \mid x_1, \dots, x_{t-1})$, based on the previous symbol $x_{t-1}$ and a recurrent context vector $\ve{h}_{t-1}$. A sample is drawn from this distribution using the Gumbel-softmax trick, yielding a soft one-hot vector that is passed through a straight-through estimator \cite{bengio2013estimating} and mapped to a constellation symbol $x_t$, following the sampling approach in \cite{stark2019joint}.
Figure~\ref{fig:encoder_blockdiagram} illustrates the recurrent operation at time step $t$ for the NPS encoder architecture. This step is repeated $L$ times to sample the sequence $\ve{x}$ from the joint distribution. 
While the encoder generates the conditional distributions, we apply the scaling 
\begin{equation}
\frac{1}{L} \sum_{t=1}^{L} \sum_{i=1}^M p(x_t = c_i) |c_i|^2 = 1
\end{equation}
as a proxy to the unit power constraint in \eqref{eq:constraint}, where the marginal symbol probabilities $p(x_t)$ are obtained from Monte Carlo estimation. 

Gumbel-softmax sampling in NPS preserves differentiability during joint-distribution learning. For practical deployment, we propose replacing it with arithmetic distribution matching \cite{baur2015arithmetic}, which maps information bits to output symbols based on the learned conditional distributions, similar to context-based source coding with adaptive arithmetic coding \cite{marpe2003context}.

\section{Channel Model}
 \label{sec:channel}
To account for channel memory inherent in optical fiber transmission, multiple symbol sequences $\ve{x}$ are generated using the encoder and concatenated before transmission. The combined sequence is passed through a channel modeled by the additive-multiplicative formulation of the first-order perturbative model as \cite{askari2024probabilistic}
\vspace{-3mm}
\begin{equation}
\label{eq:channel}
y_t = x_t \exp\left( \j \gamma \sum_{n} (|x_{t-n}|^2 - 1) c_n \right) + \Delta x_t + n_t,
\end{equation}
where $\gamma$ is the fiber nonlinearity coefficient, $c_n$ denotes the perturbation coefficient \cite{6964065} associated with the nonlinear phase rotation, $\Delta x_t$ represents the additive NLIN arising from signal-signal interactions, and $n_t$ is the lumped amplified spontaneous emission (ASE) noise at time $t$. We use this perturbative channel model for its simplicity, which speeds up training. More accurate models can also be applied. 

\section{Decoder and Loss Function}
\label{sec:decoder}
At the receiver, we discard symbols from the beginning and end of the transmission window to fully capture channel memory effects. The remaining samples are then processed by a mismatched Gaussian demapper, where the output LLRs are used to compute the binary cross entropy (BCE) loss summed across the $L$ received symbols as
\begin{equation}
    \label{eq:loss}
    \mathcal{L} = \sum_{t=1}^{L} \sum_{i=1}^{m} \mathbb{E} \left[ -\log \tilde{p}(b_t^{(i)} \mid y_t) \right],
\end{equation}
where $\mathbb{E}$ indicates statistical expectation and $\tilde{p}(b_t^{(i)} \mid y_t)$ denotes the mismatched Gaussian posterior. The loss \eqref{eq:loss} can be decomposed as
\vspace{-3mm}
\begin{equation}
\label{eq:decomposed_loss}
\begin{split}
     &\mathcal{L} = \sum_{t=1}^{L} \Big( H(\ve{b}_t) - R_t \\
    &\quad + \sum_{i=1}^{m} \mathbb{E} \left[ \Dkl(p(b_t^{(i)} \mid \ve{y}) \,\|\, \tilde{p}(b_t^{(i)} \mid y_t)) \right] \Big),
\end{split}
\end{equation}
where $\Dkl$ is the Kullback–Leibler (KL) divergence. For training, we adopt the adjusted loss $\mathcal{\hat L} = \mathcal{L} - \sum_{t=1}^{L} H(\ve{ b}_t)$, which encourages maximization of AIR, consistent with \cite{ait2020joint, stark2019joint}. Although the KL divergence between the true posterior and the mismatched Gaussian approximation in \eqref{eq:decomposed_loss} could be minimized using a learned demapper, we intentionally fix the demapper in this work to isolate the effects of transmitter-side distribution learning from receiver-side compensation.

\section{Numerical Results}
We adopt the simulation setup from \cite{askari2023probabilistic, gultekin2022kurtosis}, considering a dual-polarization $64$-QAM wavelength division multiplexing (WDM) system with $5$ channels, a baud rate of $50$~GBaud, and $55$~GHz channel spacing. Symbols are pulse shaped using a root raised-cosine filter with a $0.1$ roll-off factor. The fiber link is simulated using the split-step Fourier method (SSFM) and consists of a single span of $205$~km standard single-mode fiber with attenuation coefficient of $0.2$~dB/km, chromatic dispersion of $17$~ps/nm/km, and nonlinearity coefficient of $1.3$~W${}^{-1}$km${}^{-1}$. At the receiver, an Erbium-doped fiber amplifier (EDFA) with noise figure $5$~dB is used. Chromatic dispersion is electronically compensated, and a linear pilot-aided carrier phase recovery with a pilot rate of $2.5\%$ is applied \cite{neshaastegaran2019log}. The performance metrics for the central channel are reported.

\begin{figure}[t!]
    \centering
    \includegraphics[width=\linewidth]{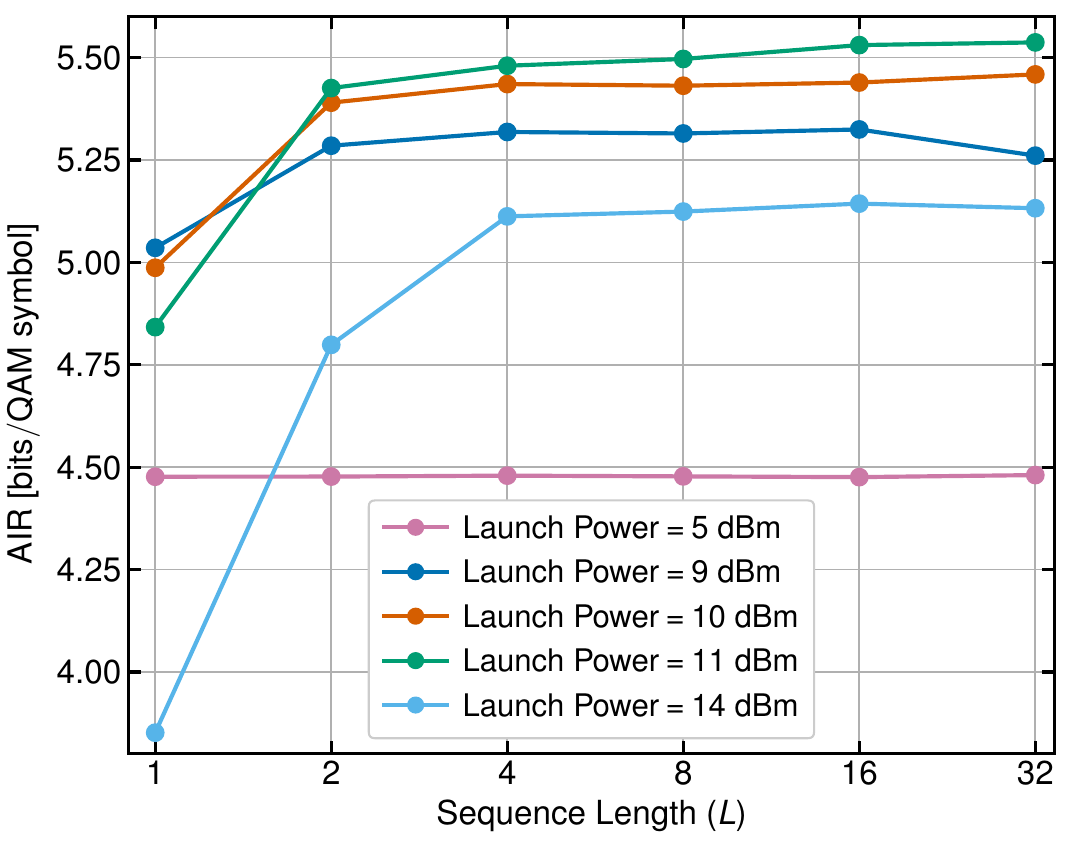}
    \caption{AIR vs. sequence length for different launch powers during training and the 
    perturbative channel model.}
    \label{fig:air_vs_l}
\end{figure}

\textsl{NPS encoder training:} We train the NPS encoder with output sequence length $L$ for each launch power using backpropagation through a single-channel, single-polarization perturbation model as described in \eqref{eq:channel}. The recurrent unit comprises a single-layer long short-term memory (LSTM) network \cite{hochreiter1997long} with a hidden size of $256$, and a linear layer that maps the hidden states to the constellation space of size $M=64$. Fig.~\ref{fig:air_vs_l} shows the AIR as a function of sequence length $L$ for models trained at different launch powers under this perturbative model. In the low power regime, joint-distribution learning (i.e., $L>1$) offers little benefit because the channel behaves approximately linearly. However, at higher launch powers, learning the joint distribution significantly improves AIR compared to marginal-distribution shaping (i.e., $L=1$). Moreover, the optimal launch power shifts from $9$~dBm at $L=1$ to $11$~dBm at $L=32$, indicating that joint-distribution learning enhances NLIN tolerance.

\begin{figure}[t!]
    \centering
    \includegraphics[width=\linewidth]{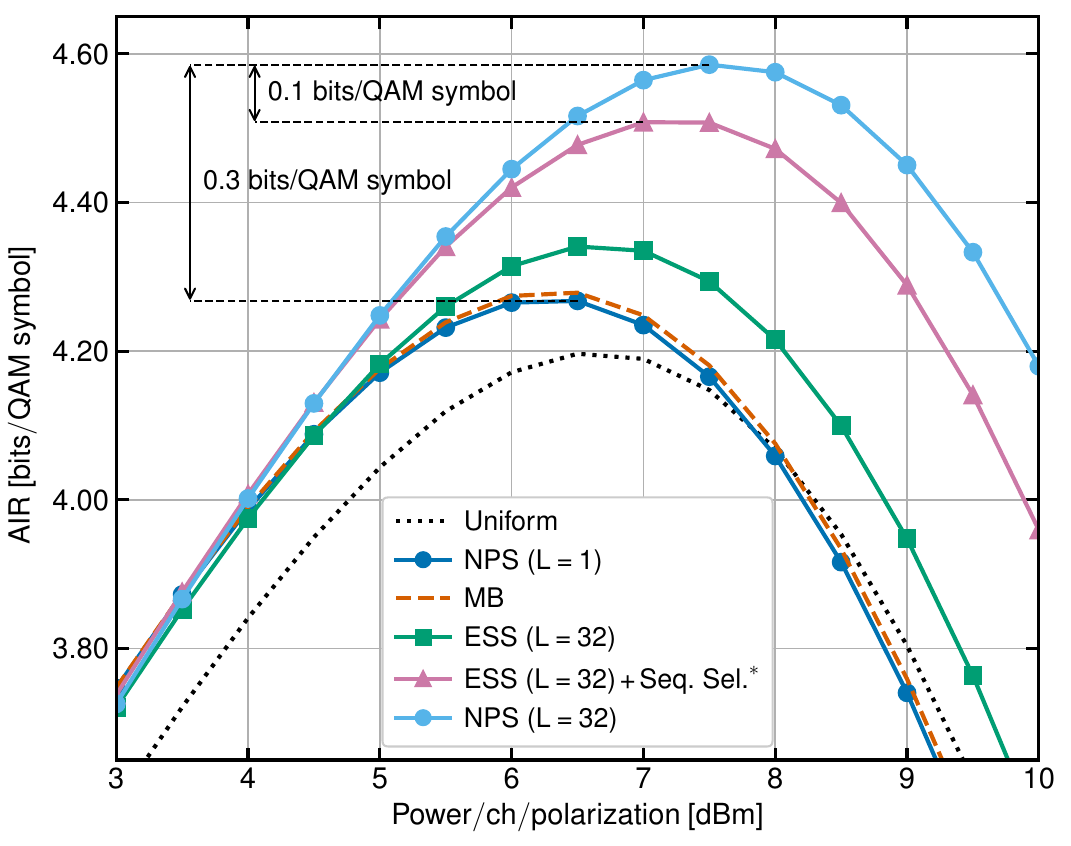}
    \caption{AIR vs. launch power for different sequence lengths. A dual-polarization WDM channel is simulated using SSFM. }
    \label{fig:air_vs_power}
\end{figure}

\textsl{SSFM simulations:} Next, we evaluate the performance of neural encoders trained with $L=1$ and $L=32$ at their respective optimal launch powers by generating sequences and testing them across a range of launch powers. Figure~\ref{fig:air_vs_power} shows the AIR versus launch power for the dual-polarization WDM channel simulated using SSFM. The AIR for uniform distribution is included as a baseline. We  compare against transmission with a Maxwell-Boltzmann (MB) distribution matched to the entropy of the NPS encoder with $L=1$, ESS with blocklength 32, and ESS with sequence selection with $64$ candidate sequences using the additive-multiplicative metric from \cite{askari2024perturbation}. The ESS rate loss due to finite-length shaping is compensated to have a fair comparison. Furthermore, we did not deduct the sequence-selection rate loss to highlight the advantage of directly learning the joint distribution by NPS. Thus, we show an upper performance bound for sequence selection.

NPS with  $L=1$ matches the performance of the MB distribution, outperforming the uniform case for most launch powers. ESS and ESS with sequence selection provide further significant improvements.
However, NPS with $L=32$ outperforms all other methods including the upper bound for ESS with sequence selection.  
At optimum launch power, the AIR gains are  0.3~bits/2D over optimized marginal PS and 0.1~bits/2D over the sequence selection bound. 
Additionally, the optimal launch power is largest for 
NPS with $L=32$,  
demonstrating its improved robustness to NLIN.

\section{Conclusions}
We proposed neural probabilistic shaping for nonlinear optical fiber channels. NPS employs an RNN with Gumbel-softmax sampling to directly model the joint symbol distribution, capturing temporal dependencies for improved performance over marginal-distribution shaping. Unlike sequence selection, it avoids additional rate loss and reduces complexity. SSFM simulations demonstrate substantial AIR gains and enhanced NLIN robustness, underscoring the effectiveness of neural sequence modeling for PS in optical systems.
\clearpage
\section{Acknowledgements}
This work was supported by Nokia Bell Labs, France; the Mitacs Accelerate International program; the Institute for Computing, Information and Cognitive Systems (ICICS) at UBC; and the Digital Research Alliance of Canada (www.alliancecan.ca).



\defbibnote{myprenote}{%
}
\printbibliography[prenote=myprenote]

\vspace{-4mm}

\end{document}